\documentclass{jmlr}

\usepackage{graphicx}
\graphicspath{ {./} }
\usepackage{multirow}
\usepackage{verbatim}

\usepackage[capposition=top]{floatrow}

\jmlrvolume{1}
\jmlryear{2018}
\jmlrworkshop{ICML 2018 AutoML Workshop}

\title[Preprocessor Selection]{Preprocessor Selection for Machine Learning Pipelines}

\author{\Name{Brandon Schoenfeld} \Email{bjschoenfeld@byu.edu} \\
    \Name{Christophe Giraud-Carrier} \Email{cgc@cs.byu.edu} \\
    \Name{Mason Poggemann} \Email{mpoggemann96@gmail.com} \\
    \Name{Jarom Christensen} \Email{jaromchristensen5@gmail.com} \\
    \Name{Kevin Seppi} \Email{kseppi@cs.byu.edu} \\
    \addr Department of Computer Science, Brigham Young University
}

\begin{document}

\maketitle

\begin{abstract}
Much of the work in metalearning has focused on classifier selection, combined more recently with hyperparameter optimization, with little concern for data preprocessing.
Yet, it is generally well accepted that machine learning applications require not only model building, but also data preprocessing.
In other words, practical solutions consist of pipelines of machine learning operators rather than single algorithms.
Interestingly, our experiments suggest that, on average, data preprocessing hinders accuracy, while the best performing pipelines do actually make use of preprocessors.
Here, we conduct an extensive empirical study over a wide range of learning algorithms and preprocessors, and use metalearning to determine when one should make use of preprocessors in ML pipeline design.
\end{abstract}


\section{Introduction}

Over the past couple of decades the work on automating the machine learning process, commonly referred to as AutoML, has mainly focused on algorithm selection~\citep{brazdiletal09}.
In recent years, this work has been extended to include the important issue of hyperparameter optimization~\citep{NIPS2015_5872,kotthoffetal16}.
Given that practical applications of machine learning often rely not only on algorithms that can be applied to data, but also on transforming the data prior to such application, the natural next step is to consider the need to design and/or select pipelines consisting of both preprocessing algorithms and model building algorithms, as in~\citep{gijsbersetal17}.

In the context of AutoML, metalearning consists of using machine learning to determine solutions to new machine learning problems based on data from the application of machine learning to other problems.
For each application of machine learning, we observe a dataset, an algorithm or sequence of algorithms, and some measure of performance.
Together, these components represent one training data point for metalearning.
With enough such data points, a metadataset can be assembled and machine learning be applied to it, with the goal of gaining insight into the machine learning process.
We extend the work of metalearning to include the selection of a single preprocessor to improve the performance of a chosen classifier, thus taking a step toward an entire ML pipeline recommendation systems.

Using data from over 10,000 machine learning experiments, we find that preprocessing data before passing it to a classification algorithm tends to hurt classification accuracy on test data, but shortens both the average training runtime and average prediction runtime for the entire pipeline.
At the same time, the most accurate pipelines (both on training and test data) within our experiments are far more likely to use some kind of preprocessing algorithm.
These results suggest that when building machine learning pipelines, especially when doing so automatically, preprocessors should be used, but must be chosen carefully.
We make use of metalearning on the data we collected to build predictive models of when a preprocessing algorithm will improve a particular classifier's accuracy or runtime.
Our results suggest that metalearning can intelligently reduce the search space for AutoML systems by guiding the preprocessor selection process when building entire machine learning pipelines.

\section{Experimental Setup}

We select 192 classification datasets from OpenML~\citep{van14}.
We use approximately a 70/30 train/test for each dataset, since cross-validation would result in prohibitive computational cost for data collection.
Since many preprocessing and classification algorithms do not tolerate sparse or non-numeric data, especially as implemented in scikit-learn~\citep{pedregosaetal11}, we ``clean'' all datasets by first imputing missing values, and second one-hot encoding all categorical variables.
Performing this cleaning on all datasets allows for controlled comparisons across experiments.
Any differences in pipeline performance between two experiments cannot be attributed to this cleaning as it is held constant across all experiments.

Imputation is performed on each variable, independent of the others, and agnostic to the target class, by randomly selecting a known value from that same variable.
This imputation method assumes that missing values are not a separate category of their own, but that they represent data that was either not measured, not recorded, or lost.
This method has the advantage over imputing the mean or the mode in that it naturally tends to preserve the original distribution of data in each column.

We consider all possible pipelines of length 3 or 4, with the first 2 steps fixed to our data cleaning process, i.e., imputer and encoder, followed by 0 or 1 preprocessor, and finally 1 classifier, for a total of 10,368 pipelines.
We select 8 preprocessing algorithms and 6 classification algorithms, all implemented in scikit-learn, as follows:
\begin{itemize}
    \item \emph{Preprocessing Algorithms}: Min-Max Scaler (MMS), Standard Scaler (SS), Select Percentile (SP), Principal Component Analysis (PCA), Fast Independent Component Analysis (ICA), Feature Agglomeration (FA), Polynomial Features (PF), and Radial Basis Function Sampler (RBFS)
    \item \emph{Classification Algorithms}: Random Forest Classifier (RFC), Logistic Regression (LR), K-nearest Neighbors Classifier (KNN), Perceptron (Per), Support Vector Classifier (SVC), and Gaussian Naive Bayes (GNB)
\end{itemize}
All algorithms use their default hyper-parameter settings.
Since our interest lies in determining which preprocessing algorithms to use, if any, we setup, as a \emph{baseline} comparison, the 1,152 pipelines of length 3, which do not use any preprocessing algorithm.
We only compare pipelines of length 4, i.e. with a preprocessor, against the baseline pipeline run on the same dataset, such that the only difference between compared pipelines is whether a preprocessor is present.

\section{Base Experiment Results}

Of the 10,368 possible pipelines, 10,331 ran to completion.
Thirty pipelines failed due to a known, but yet unresolved, bug in the eigenvalue decomposition algorithm used by ICA and another one because ICA did not converge with default hyperparameters. 
The final six failed on one dataset because the combination of one-hot encoding and the PF preprocessor caused the dimensionality of the feature space to explode, and all the classifiers ran out of memory.
We chose not to adjust the hyperparameters or modify the dataset to maintain comparability with the other pipelines.

\begin{table}[t]
    \centering
    \setlength\tabcolsep{4pt} 
    \begin{tabular}{|c||c|c|c|c|c|c||c|} \hline
          & RFC       & LR        & KNN       & Per         & SVC       & GNB       & Mean      \\ \hline \hline
    MMS   & 192,\textbf{23},23 & 192,\textbf{43},43 & 191,\textbf{84},83 & 191,\textbf{104},104 & 159,\textbf{77},68 & 110,\textbf{42},27 & 173,\textbf{62},58 \\ \hline
    SS    & 192,\textbf{58},58 & 181,\textbf{89},84 & 188,\textbf{98},96 & 192,\textbf{117},117 & 138,\textbf{93},71 & 101,\textbf{81},41 & 165,\textbf{89},78 \\ \hline
    SP    & 192,\textbf{27},27 & 192,\textbf{45},45 & 192,\textbf{65},65 & 192,\textbf{56},56   & 176,\textbf{60},55 & 152,\textbf{84},71 & 183,\textbf{56},53 \\ \hline
    PCA   & 173,\textbf{59},58 & 172,\textbf{58},50 & 192,\textbf{30},30 & 192,\textbf{88},88   & 155,\textbf{4},4   & 90,\textbf{106},52 & 162,\textbf{57},47 \\ \hline
    ICA   & 161,\textbf{38},38 & 186,\textbf{56},56 & 187,\textbf{77},77 & 186,\textbf{83},83   & 110,\textbf{31},20 & 99,\textbf{75},39  & 155,\textbf{60},52 \\ \hline
    FA    & 191,\textbf{13},13 & 192,\textbf{30},30 & 192,\textbf{25},25 & 191,\textbf{27},26   & 165,\textbf{33},31 & 155,\textbf{64},52 & 181,\textbf{32},30 \\ \hline
    PF    & 140,\textbf{97},71 & 42,\textbf{96},13  & 77,\textbf{49},21  & 84,\textbf{79},29    & 32,\textbf{45},17  & 9,\textbf{71},5    & 64,\textbf{72},26  \\ \hline
    RBFS  & 119,\textbf{23},17 & 172,\textbf{32},24 & 165,\textbf{21},18 & 181,\textbf{59},53   & 86,\textbf{22},7   & 41,\textbf{66},11  & 127,\textbf{37},22 \\ \hline \hline
    Mean & 170,\textbf{42},38 & 166,\textbf{56},43 & 173,\textbf{56},52 & 176,\textbf{76},70   & 128,\textbf{45},34 & 95,\textbf{73},37  & 151,\textbf{58},46 \\ \hline
    \end{tabular}
    \caption{Number of Tasks for which Preprocessors Improved upon Baseline}
    \floatfoot{\footnotesize{The comma separated values in each cell reflect the number of datasets for which the preprocessor improved train time, \textbf{test accuracy}, and both, in order, compared to the baseline pipeline. Preprocessors are enumerated along the rows and classifiers along the columns.}}
    \label{tab:whichpreproc}
\end{table}

\subsection{Runtime}

An unexpected and counter-intuitive result is that adding a preprocessing step in a pipeline can actually reduce a pipeline's total runtime.
79.1\% of the 9,179 non-baseline pipelines were trained in less time than their baseline.
Similarly, when making predictions on test data, 80.2\% of the non-baseline pipelines run in less time.

We observe in Table~\ref{tab:whichpreproc} that preprocessors which reduce the number of features in the dataset reduce runtime more often than those which create or add new features.
For example, SP, which reduces the dimensionality by 90\%, and FA, which keeps only two features, almost always reduce training time, while PF, which adds a quadratic number of features, more often increases training time.
Since the algorithmic complexity of the classifiers often depends on the number of features, feature reduction will reduce runtime for the classifier.
Based on these results, it appears that the cost of training a preprocessor is more than paid by the reduction in classifier training time.

\begin{figure}[t]
    \centering
    \includegraphics[trim={.5cm .5cm .5cm .5cm}, width=0.49\textwidth]{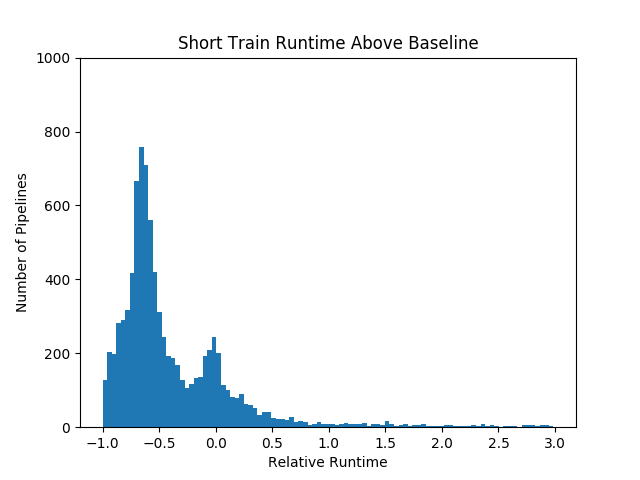}
    \includegraphics[trim={.5cm .5cm .5cm .5cm}, width=0.49\textwidth]{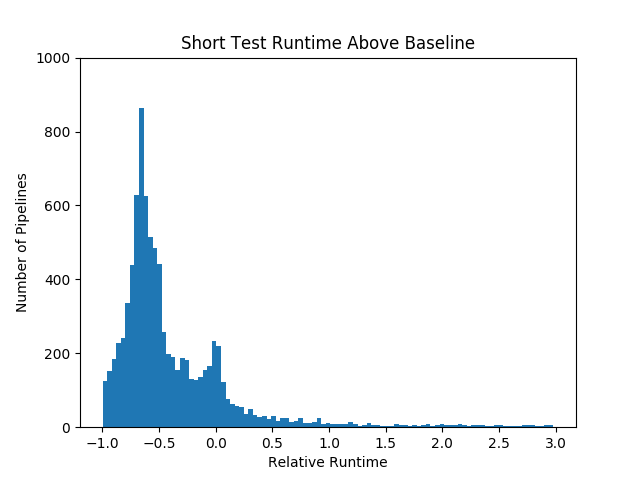}
    \caption{Relative Pipeline Runtime Above Baseline}
    \floatfoot{\footnotesize{Both graphs have long tails which we have truncated at 3.0. The train graph (left) displays 8,764 of the 9,719 non-baseline pipelines; the maximum relative runtime is 2,024. The test graph (right) displays 8,709 pipelines, with a maximum relative runtime of 1,041. A value of 1.0 indicates that a pipeline took 100\% more time to train or test than the baseline, while -0.5 means a pipeline took 50\% less time.}}
    \label{fig:piptime}
\end{figure}

Figure~\ref{fig:piptime} shows the relative training time of pipelines, compared to their baseline.
Interestingly, 25\% are at least 67.9\% faster than baseline.
While most of the mass in the plots of Figure~\ref{fig:piptime} lies below zero, there is a long tail with 3 pipelines taking over 1,000 times longer to train than their baseline because of the presence of the PF preprocessor.

\subsection{Accuracy}
Unlike runtime, preprocessing data tends to hinder pipeline accuracy.
Of the 9,179 non-baseline pipelines, 69.4\% yielded lower test accuracy than the baseline.
However, considering the most accurate of all 10,331 pipelines for each dataset, we find that 91.1\% do use a preprocessor.
Similarly, for pipelines with accuracy within 5\% and 10\% of the top, 85.7\% and 86.4\%, respectively, do use a preprocessor.

Table~\ref{tab:whichpreproc} shows that SS improves test accuracy for more pipelines than any other preprocessor.
Furthermore, Table~\ref{tab:preprocacc} shows that SS is the only preprocessor which improves both pipeline train and test accuracy on average.
Perhaps classifiers can optimize over the feature space better when the data is zero-centered with unit variance.
A similar argument could be made for MMS, which is the only other preprocessor to show any improvement in test accuracy.
However, it might be the case that these preprocessors are simply well-suited to the datasets we considered.

\begin{table}[b]
    \centering
    \begin{tabular}{|c||c|c|c|c|} \hline
    Preprocessor & MMS & SS & SP & PCA \\ \hline
    Mean & -0.3\%, 0.8\% & 2.3\%, 2.3\% & -10.0\%, -5.71\% & 0.2\%, -0.4\% \\ \hline
    StdDev & 14.2\%, 15.8\% & 13.1\%, 15.7\% & 19.4\%, 18.9\% & 11.8\%, 12.4\% \\ \hline \hline
    Preprocessor & ICA & FA & PF & RBFS \\ \hline
    Mean & -6.2\%, -6.2\% & -15.9\%, -15.5\% & 0.7\%, -1.9\% & -8.3\%, -15.8\% \\ \hline
    StdDev & 24.6\%, 22.2\% & 23.5\%, 25.7\% & 13.1\%, 13.6\% & 24.2\%, 24.9\% \\ \hline
    \end{tabular}
    \caption{Preprocessor Breakdown of Pipeline Accuracy Above Baseline}
    \floatfoot{\footnotesize{In each cell, the first number is for training data and the second one for test data.}}
    \label{tab:preprocacc}
\end{table}

\subsection{Combined Accuracy and Runtime}

Given the above results, one may wonder if it is possible to get both improved accuracy and runtime or if there exists a natural trade-off between the two.
Thus, we turn our attention to the combined effect of accuracy on runtime, and vice-versa. By focusing on the high end of each spectrum, we observe the following.
\begin{itemize}
    \item For the 25\% fastest training non-baseline pipelines, the average accuracy is 9.2\% lower than would be obtained without the preprocessor.
    \item For the 25\% fastest testing non-baseline pipelines, the average accuracy is 7.8\% lower than would be obtained without the preprocessor.
    \item For the 25\% most accurate non-baseline pipelines, the average training time is 126.6\% higher, and the average test time is 93.1\% higher, than would be obtained without the preprocessor.
\end{itemize}
On average, it appears that there is indeed no free lunch: a preprocessor will increase accuracy at the cost of runtime, or it will reduce runtime at the cost of accuracy.
It thus becomes important to be able to intelligently select preprocessors so as to reduce the likelihood of choosing the wrong preprocessor.

\section{Meta Experiment Results}
\label{sec:metaexp}

We turn our attention to metalearning to further investigate the details of the above relationships, and thus inform the AutoML process.
The predictive features of our metadataset consist of metafeatures extracted from the base dataset and a one-hot encoded preprocessor identifier.
We extract 18 simple, 8 statistical, 1 information-theoretic, and 14 landmarking metafeatures, from each of the 192 base datasets.
Our metafeature extraction tool is an extension of an existing R script~\citep{reif}.
The target class is a binary value indicating whether or not the given preprocessor produces a pipeline with test accuracy greater than or equal to the corresponding baseline pipeline.
We use the random forest classifier (RFC) as our metalearner, as it showed best empirical performance on the meta experiments.

The accuracy of the metamodel at predicting whether a given preprocessor (passed as input as part of the predictive features) will improve accuracy over not using some preprocessor is 62.6\% on the test data.
In contrast, if we instead merely select the mode class (which happens to be always choosing that the preprocessor will not help the classifier) we obtain 53.0\% accuracy.

\section{AutoML with Metalearning}

The value of metalearning in the context of AutoML systems is found when trained metamodels can be used to design better pipelines than alternative strategies.
Here, we simulate an AutoML scenario in which a system must decide which preprocessor to use, given a particular task dataset and given that it has already chosen a classifier.
To assess the added value of our metamodel, we create four agents which make this decision, one of which uses the RFC metamodel. Their strategies are:
\begin{enumerate}
    \item \emph{None Agent}. Choose never to preprocess. This represents much of the state-of-the-art, where only a classifier is selected.
    \item \emph{Random Agent}. Choose a preprocessor, or none, uniformly at random. This represents the case where the agent has no special prior knowledge of the preprocessors.
    \item \emph{Mode Agent}. Choose the preprocessor that is empirically the best, for each base classifier, based on past experiments. A veteran data scientist might begin here.
    \item \emph{Oracle Agent}. From the subset of preprocessors that the RFC metamodel predicts would be good, choose the one that is empirically the best based on past experiments.
\end{enumerate}

The AutoML simulator splits the entire metadataset, of all experiment results generated earlier, into a 70/30 train/test split.
Each agent is allowed to use the training portion of the metadata to inform their decisions on the test portion, though the None and Random agents do not.
At test time, each agent is given a particular dataset and classifier and must decide which preprocessor to use, if any.

The Oracle agent induces a metamodel for each base classifier, as in section~\ref{sec:metaexp}.
It queries the appropriate metamodel for each of the eight preprocessors. Of the preprocessors predicted to help, it selects the one that helped most often in the training metadata.
If none of the preprocessors were predicted to help, none is chosen.
Note that the Mode agent is a metalearning agent which takes full advantage of the training metadata, giving it a tremendous advantage over the None and Random agents.

We measure the performance of each agent by computing what percent worse each agent is than an \emph{Optimal Agent} who knows exactly which preprocessor (or none) to use, on each test task.
This is only achieved by brute force search: running all possible pipelines.
We aggregate these results across all test tasks, as shown in Table~\ref{tab:agentperf}.
The two metalearning agents, Mode and Oracle, perform best.

\begin{table}[b]
    \centering
    \begin{tabular}{|c||c|c|c|c|c|} \hline
    Agent  & None     & Random   & Mode    & Oracle  & Optimal \\ \hline
    Mean   & -10.51\% & -15.40\% & -6.57\% & -6.19\% & 0       \\ \hline
    StdDev & 17.23\%  & 21.41\%  & 12.23\% & 12.25\% & 0       \\ \hline
    \end{tabular}
    \caption{Agent Comparison in AutoML Simulation}
    \label{tab:agentperf}
    \floatfoot{\footnotesize{How much worse the agent-chosen pipelines performed relative to the best known pipelines for the test task.}}

\end{table}

\section{Conclusion}

We have analyzed the impact of selecting from among eight preprocessors in the context of six classification learning algorithms for the design of short ML pipelines.
Our results suggest that preprocessing does not always improve accuracy, but in many cases does reduce training and prediction time.
Using metalearning, we have shown that a metamodel can help AutoML systems determine more intelligently which preprocessor, if any, might help improve a pipeline's performance.
Choosing the best metamodel or even metapipeline is left to future work.
Future work also includes exploring longer pipelines, including hyper-parameter selection, and agents considering pipeline execution times.

\acks{
    We gratefully acknowledge support from the Defense Advanced Research
    Projects Agency under award FA8750-17-2-0082.
}

\bibliography{refs}

\end{document}